\documentclass{article}

\usepackage{microtype}
\usepackage{graphicx}
\usepackage{subfigure}
\usepackage{subcaption}
\usepackage{booktabs} 

\usepackage{hyperref}

\usepackage[accepted]{icml2024}

\usepackage{amsmath}
\usepackage{amssymb}
\usepackage{mathtools}
\usepackage{amsthm}

\usepackage[capitalize,noabbrev]{cleveref}

\theoremstyle{plain}

\theoremstyle{definition}

\theoremstyle{remark}

\usepackage[textsize=tiny]{todonotes}

\icmltitlerunning{FairPFN: Transformers Can do Counterfactual Fairness}

\begin{document}

\twocolumn[
\icmltitle{FairPFN: Transformers Can do Counterfactual Fairness}



\icmlsetsymbol{equal}{*}

\begin{icmlauthorlist}
\icmlauthor{Jake Robertson}{yyy,ddd}
\icmlauthor{Noah Hollmann}{xxx}
\icmlauthor{Noor Awad}{yyy}
\icmlauthor{Frank Hutter}{yyy,zzz}
\end{icmlauthorlist}

\icmlaffiliation{yyy}{University of Freiburg, Freiburg, Germany}
\icmlaffiliation{xxx}{Charité, Berlin, Germany}
\icmlaffiliation{zzz}{ELLIS Institute Tübingen, Tübingen, Germany}
\icmlaffiliation{ddd}{Zuse School ELIZA, Darmstadt, Germany}


\icmlcorrespondingauthor{Jake Robertson}{robertsj@cs.uni-freiburg.de}

\icmlkeywords{Machine Learning, ICML}

\vskip 0.3in
]



\printAffiliationsAndNotice{} 

\begin{abstract}



Machine Learning systems are increasingly prevalent across healthcare, law enforcement, and finance but often operate on historical data, which may carry biases against certain demographic groups. 
Causal and counterfactual fairness provides an intuitive way to define fairness that closely aligns with legal standards. 
Despite its theoretical benefits, counterfactual fairness comes with several practical limitations, largely related to the reliance on domain knowledge and approximate causal discovery techniques in constructing a causal model. In this study, we take a fresh perspective on counterfactually fair prediction, building upon recent work in in-context-learning (ICL) and prior-fitted networks (PFNs) to learn a transformer called FairPFN. This model is pre-trained using synthetic fairness data to eliminate the causal effects of protected attributes directly from observational data, removing the requirement of access to the correct causal model in practice.
In our experiments, we thoroughly assess the effectiveness of FairPFN in eliminating the causal impact of protected attributes on a series of synthetic case studies and real-world datasets. Our findings pave the way for a new and promising research area: transformers for causal and counterfactual fairness.
\end{abstract}

\section{Introduction}

Algorithmic bias is one of the most pressing AI-related risks, arising when ML-assisted decisions produce discriminatory outcomes towards historically underprivileged demographic groups \cite{angwin-propub16}. 
Despite the topic of fairness receiving significant attention in the ML community, various critics from outside the fairness community argue that statistical measures of fairness and current methods to optimize them are largely misguided in terms of their context-dependence and transferability to effective legislation. 
Recent work in causal fairness has proposed the popular notion of counterfactual fairness, which provides the intuition that outcomes are the same in the real world as in the counterfactual world where \emph{protected attributes} - such as gender, ethnicity, or sexual orientation - take on a different value. 
According to a recent review contrasting observational and causal fairness metrics \cite{castelnovo2022clarification}, the non-identifiability of causal models from observational data \citep{peters2012identifiability} presents a significant challenge in applying causal fairness in practice, as causal mechanisms are often complex due to the intricate nature of bias in real-world datasets.
If causal model assumptions are incorrect - for example, when a covariate is assumed not to be influenced by a protected attribute when in fact it is - proposing the wrong causal graph can provide a false sense of security and trust \cite{ma2023learning}. 

In this study, we introduce a novel approach to counterfactual fairness based on the recently proposed TabPFN. Our transformer-based approach coined FairPFN, is pre-trained on a synthetic benchmark of causally generated data and learns to identify and remove the causal effect of protected attributes. In our experimental results across a series of synthetic case-studies and real-world datasets, we demonstrate the effectiveness, flexibility, and extensibility of transformers for causal and counterfactual fairness. 

\begin{figure*}[t!]
    \centering
    \includegraphics[width=0.9\linewidth]{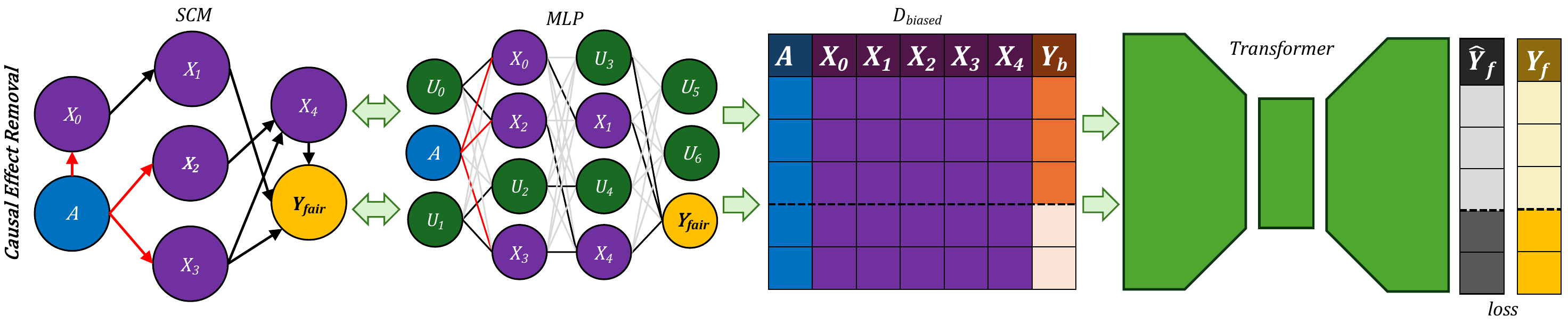}
    \caption{\textbf{FairPFN Pre-training:} FairPFN is pre-trained on a synthetic prior of datasets generated from sparse SCMs with exogenous protected attributes. A biased dataset is generated and passed as context to the transformer, and the loss is calculated with respect to the fair outcomes calculated by removing the causal influence of the protected attribute.}
    \label{fig:pretraining}
\end{figure*}

\section{Background}


\paragraph{Algorithmic Fairness} Algorithmic bias occurs when past discrimination against a demographic group such as ethnicity or sex is reflected in the training data of an ML algorithm. In such cases, ML algorithms are well known to reproduce and even amplify this bias in their predictions \cite{barocas2023fairness}. Fairness as a topic of research concerns the measurement of algorithmic bias and the development of principled methods that produce non-discriminatory predicted outcomes.


\textbf{Causal Fairness Analysis} Causal ML is a new and emerging research field that aims to represent data-generating processes and prediction problems in the language of causality, offering support for causal modeling, mediation analysis, and counterfactual explanations. 
The Causal Fairness Analysis (CFA) framework \cite{plecko2022causal} draws parallels between causal modeling and legal doctrines of direct and indirect discrimination. By categorizing variables into protected attributes $A$, mediators $X_{med}$, confounders $X_{conf}$, and outcomes $Y$, the CFA defines the Fairness Cookbook of causal fairness metrics: the Direct Effect (DE), Indirect Effect (IE), and Spurious Effect (SE). These metrics facilitate mediation analysis to assess the remaining causal effects of various bias-mitigation approaches.

\begin{figure*}[t!]
    \centering
    \includegraphics[width=0.9\linewidth]{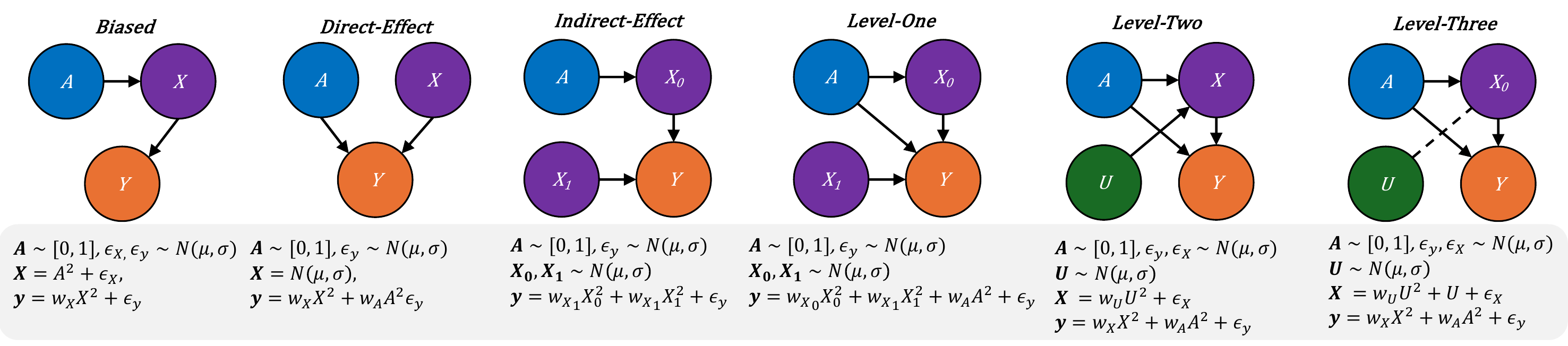}
    \caption{\textbf{Causal Case Studies:} Visualization and data generating processes of synthetic causal case studies, a handcrafted set of benchmarks designed to evaluate FairPFN's ability to remove various sources of bias in causally generated data.}
    \label{fig:casestudies}
\end{figure*}

\begin{figure}[h!]
    \centering
    \includegraphics[width=0.875\linewidth]{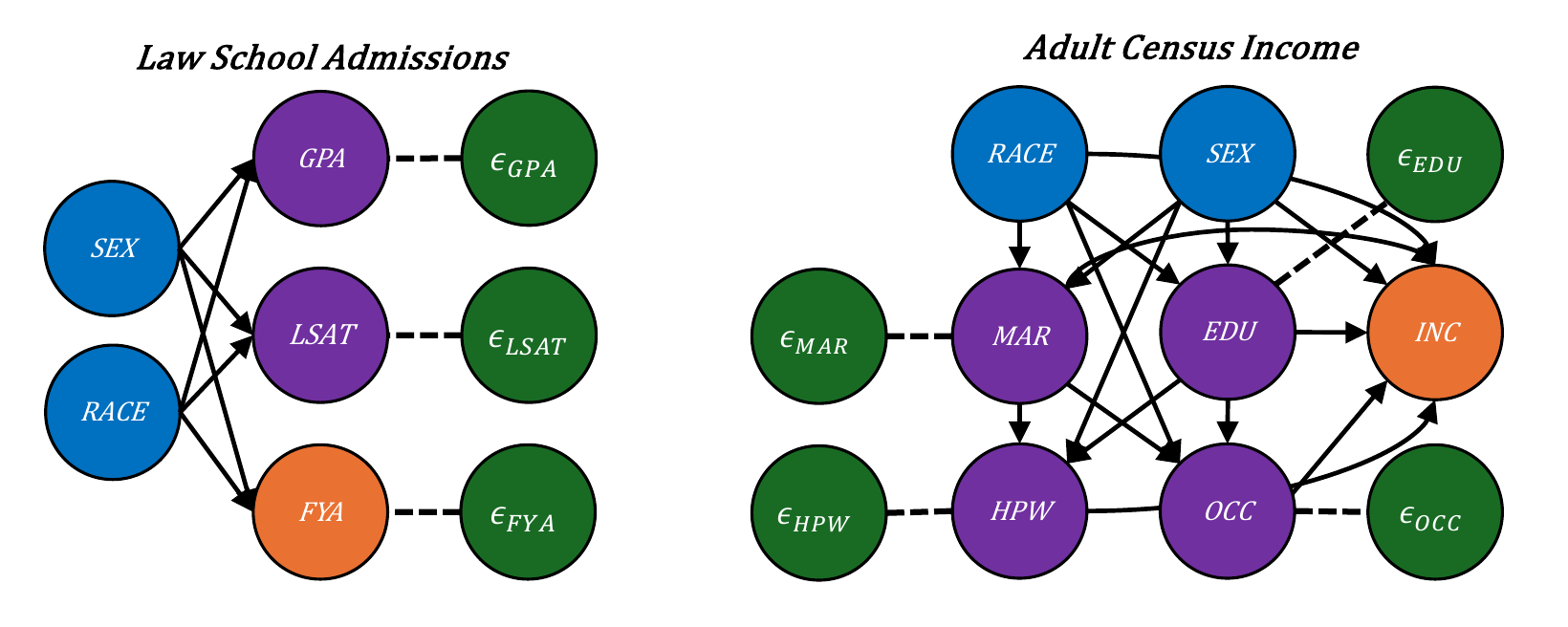}
    \caption{\textbf{Real-World Datasets}: Causal graphs of real-world datasets Law School Admissions and Adult Census Income.}
    \label{fig:real-world}
\end{figure}

\textbf{Counterfactual Fairness} 
A related causal concept of fairness is counterfactual fairness \cite{kusner2017counterfactual}, which requires that outcomes remain the same in both the real world and a counterfactual world where a protected attribute assumes a different value. Given a causal graph, counterfactual fairness can be obtained either by fitting to observable non-descendants (Level-One), the inferred values of an exogenous unobserved variable (Level-Two) or the noise terms of an Additive Noise Model for observable variables (Level-Three). Counterfactual fairness has gained significant popularity in the fairness community, inspiring recent work on path-specific extensions \cite{peters2014causal} and the application of Variational Autoencoders (VAEs) to achieve counterfactually fair latent representations\footnote{CLAIRE is not included as a baseline as their training code or model is not publicly available.} \cite{ma2023learning}.

A key challenge in the CFA, counterfactual fairness, and causal ML, in general, is the assumption regarding the prior knowledge of causal graphs and models, which relies heavily on domain knowledge and approximate causal discovery techniques. 
In the context of fairness, \cite{castelnovo2022clarification} argue that it is challenging to obtain causal graphs representing complex systemic inequalities. Additionally, \cite{ma2023learning} demonstrate that proposing an incorrect causal graph or model can deteriorate counterfactual fairness and potentially lead to adverse impacts (e.g. fairwashing) if the causal relationships between protected attributes and other variables are incorrectly assumed.

\textbf{Prior-Fitted Networks} Prior-Fitted Networks (PFNs) are a recent approach to incorporating prior knowledge into neural networks via pre-training on datasets sampled from a prior distribution \cite{muller2021transformers}. This allows PFNs to perform well on downstream tasks with limited data.

TabPFN \cite{hollmann2022tabpfn}, a recent application of PFNs to small, tabular classification problems, trains a transformer on a hypothesis of synthetic datasets generated from sparse SCMs, achieving state-of-the-art results by integrating over the simplest causal explanations for the data in a single forward pass of the network. 

\section{Methodology}

In this section, we introduce FairPFN, a novel bias mitigation technique that synergizes concepts from prior-fitted networks (PFNs) with principles of causal and counterfactual fairness. FairPFN aims to eliminate the causal and counterfactual effects of protected attributes using only observational data.

\textbf{Synthetic Prior Data Generation} The main methodological contribution of FairPFN is its fairness prior, designed to represent the causal mechanisms of bias in real-world data. FairPFN's fairness prior includes a key addition to the TabPFN hypothesis space, namely the inclusion and specification of protected attributes in the randomly generated SCMs as \textit{exogenous} variables\footnote{The simplifying assumption of exogenous protected attributions is commonly made in the causal fairness literature as protected attributes are typically unchangeable by definition and hold ancestral closure \cite{plecko2022causal}}.

The first step of FairPFN is the generation of \textit{biased} synthetic datasets that realistically represent the causal mechanisms of bias in real-world datasets. We provide a visual overview of this process in (Figure \ref{fig:pretraining}). Taking inspiration from TabPFN, we represent SCMs as Multi-Layer-Perceptrons (MLPs) with linear layers serving to represent the structural equation $f = P\cdot W^Tx + \epsilon$ where $W$ are the weights of the activations, $\epsilon$ is Gaussian Noise, and $P$ is a dropout mask sampled from a log-scale to encourage sparsity of the represented SCM. 

The exogenous protected attribute is sampled from the input to the MLP as a binary variable $A \in \{a_0, a_1\}$ where $a_i$ is sampled from the same range as non-protected exogenous variables $U_{fair}$ to prevent numeric overflow. We uniformly sample $m$ features $X$ from the second hidden layer on to ensure that they contain rich representations of the causes. Finally, we select the target $Y$ from the output layer. Because $Y$ is a continuous variable, we binarize over a random threshold. We note that without binarizing, future versions of FairPFN are extensible to regression tasks and handling multiple protected attributes. 

Via a forward pass of the MLP, we generate a dataset $D_{bias} = (A, X_{bias}, Y_{bias})$ of $n$ samples and repeat this process throughout training on randomly sampled SCMs, number of features, and number of samples to generate a rich synthetic representation of real-world, biased data.

\textbf{FairPFN Pre-training} The strategy by which we pre-train the transformer to perform counterfactual fairness is by generating two datasets, $D_{bias}$ and $D_{fair}$. The fair dataset is generated by performing dropout on the outgoing edges of the protected attribute in the sampled MLP. This has the effect of setting the weight to $0$ in the represented equation $f = 0\cdot wx + \epsilon$, meaning that the effect of the protected attribute is reduced to Gaussian noise $\epsilon$ as visualized in Figure \ref{fig:corr_removal}.\footnote{We note that this bias removal strategy motivates our sampling of $A$ from an arbitrary distribution $A\in \{a_0, a_1\}$ and not $A\in \{0, 1\}$ because $f = 0\cdot wx + \epsilon$ would have the same result as $f = p\cdot 0x + \epsilon$}. Having generated two datasets, we pass in $D_{bias}$ as context to the transformer, and calculate the loss with respect to the transformer's predictions and the fair outcomes $Y_{fair}$ (Figure \ref{fig:effect_removal}). It's worth noting that we simply discard $X_{fair}$ in this strategy, but discuss how it could be applied to train FairPFN to be a fairness pre-processsing technique in Section \ref{sec:conclusion}.

\textbf{Fairness Prior-Fitting} We train the transformer for approximately 3 days on an \texttt{RTX-2080} GPU. Throughout training, we vary several hyperparameters, including the size and connectivity of the MLPs, the number of features sampled, and the number of dataset samples generated. To calculate the loss between the predicted and ground truth values of $Y_{fair}$ classification setting, we apply Binary-Cross-Entropy (BCE) loss and a decaying learning rate schedule.

\textbf{Causal Case Studies} First, we introduce our synthetic benchmark, a hand-crafted set of causal case studies with increasing difficulty, designed to evaluate FairPFN's ability to remove various sources of bias in causally generated data. 

Our simplest case study is the \texttt{Biased} scenario, where the protected attribute $A$ has an indirect causal effect on the outcome. This case study aims to simulate what happens when FairPFN encounters a scenario where the outcome is only causally influenced by a protected attribute. Next, we implement \texttt{Direct} and  \texttt{Indirect Effect} scenarios to evaluate FairPFN's ability in isolating the direct and indirect effects of bias. Finally, we implement three scenarios,  \texttt{Level-One}, \texttt{Level-Two}, and \texttt{Level-Three} with inspiration drawn from the three levels of counterfactual fairness. We provide an overview of our causal case studies with their corresponding data-generating processes in Figure \ref{fig:casestudies}.

To provide a diverse synthetic benchmark, we independently generate 100 datasets per case study varying the causal weights of simulated protected attributes $w_A$, the number of samples $m \in (100, 1000)$ (log-scale), and the standard deviation of Gaussian noise terms $\sigma \in (0, 1)$ (log-scale). We also create counterfactual versions of each dataset which we use to evaluate FairPFN for counterfactual fairness, which we measure as the Mean-Absolute Error (MAE) between predictions on the real and counterfactual datasets.

\textbf{Real-World Datasets} We also apply FairPFN to two real-world datasets whose causal graphs are widely agreed upon in the causal fairness community. The first problem we focus on is the Law School Admissions problem, which comes from the 1998 LSAC National Longitudinal Bar Passage Study \cite{wightman1998lsac}. The LSAC study recorded law school admissions data from approximately 30,000 applicants to top US law schools and reports a significant disparity of bar passage and first-year average (FYA) outcomes with respect to applicant race.

We use the causal graph visualized in Figure \ref{fig:real-world} and observational data as input to the \texttt{dowhy.gcm} module \cite{sharma2020dowhy}, which fits a causal model using Random Forest Regressors to estimate non-linear causal relationships. We use these causal models to measure the TE and create counterfactual data. We also apply the \texttt{compute\_noise} functionality to infer the values of noise terms $\epsilon_{GPA}$ and $\epsilon_{LSAT}$ to use later as training data for our \texttt{Level-Three} baseline (Appendix Section \ref{sec:baselines}).

The next problem we focus on is the Adult Census Income problem \cite{asuncion2007uci}, a dataset drawn from the 1994 US Census that records the demographic information and income outcomes ($INC \geq 50K$) for nearly 50,000 individuals. Again, we fit a causal model in order to measure the TE of protected attribute $RACE$, create a counterfactual dataset (Figure \ref{fig:cntf}), and infer values of noise terms $\epsilon$ (Appendix Figure \ref{fig:noise}).

\section{Results}

In this section, we evaluate the performance of FairPFN on our benchmark of synthetic and real-world scenarios, with the key message that FairPFN removes the causal and counterfactual effect of protected attributes without any knowledge of the causal model. 

\begin{figure}[t!]
    \centering
    \includegraphics[width=\linewidth]{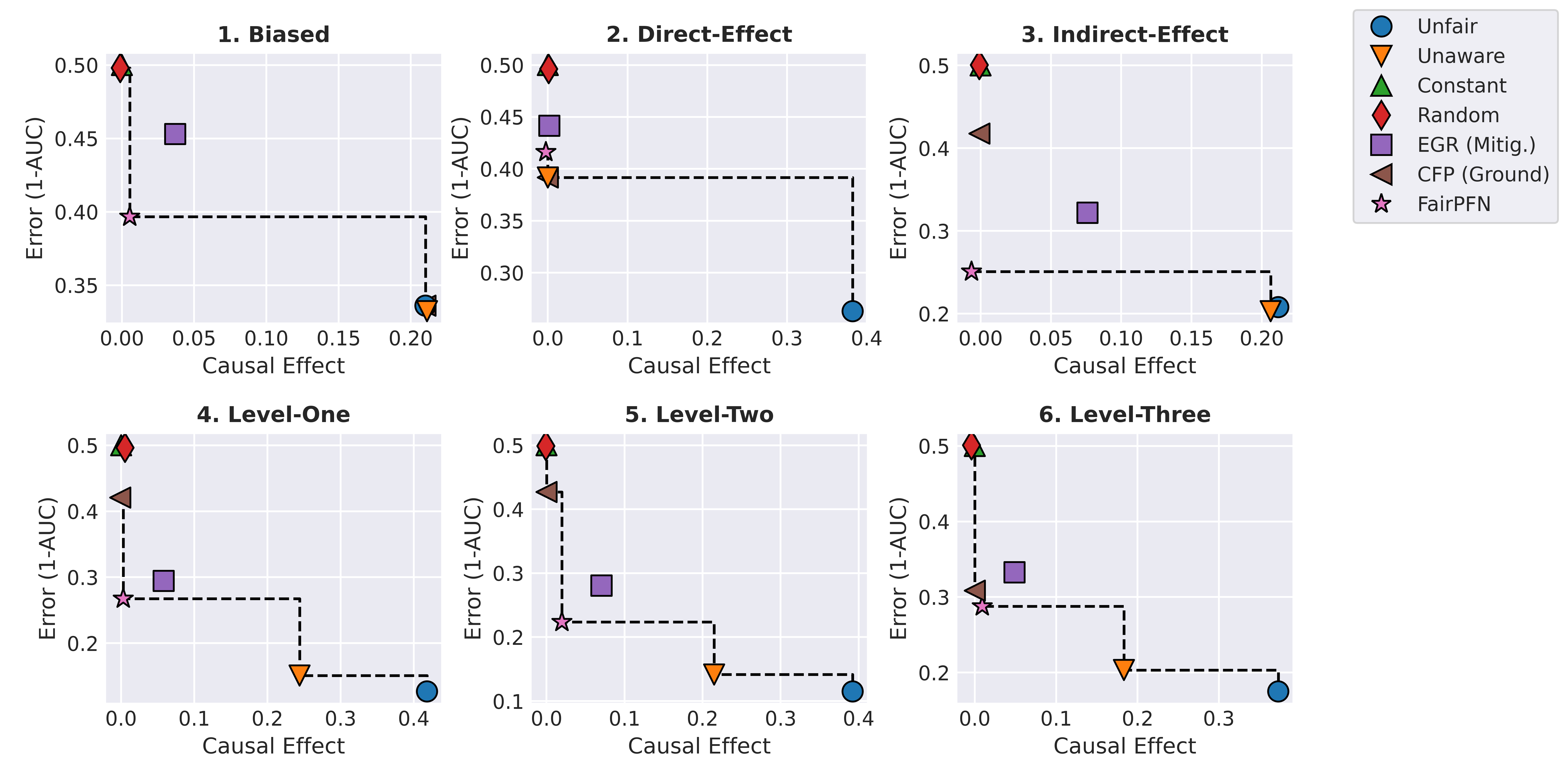}
    \caption{\textbf{Causal Effect Removal (Synthetic):} Average causal effect (IE, DE, or TEE) and error (1-AUC) of FairPFN compared to our baselines. FairPFN is on the Pareto Front across all synthetic case studies, dominates \texttt{EGR} on 5 out of 6, and always improves upon \texttt{CFP} in terms of error.}
    \label{fig:effect_removal}
\end{figure}

\begin{figure}[t!]
    \centering
    \includegraphics[width=\linewidth]{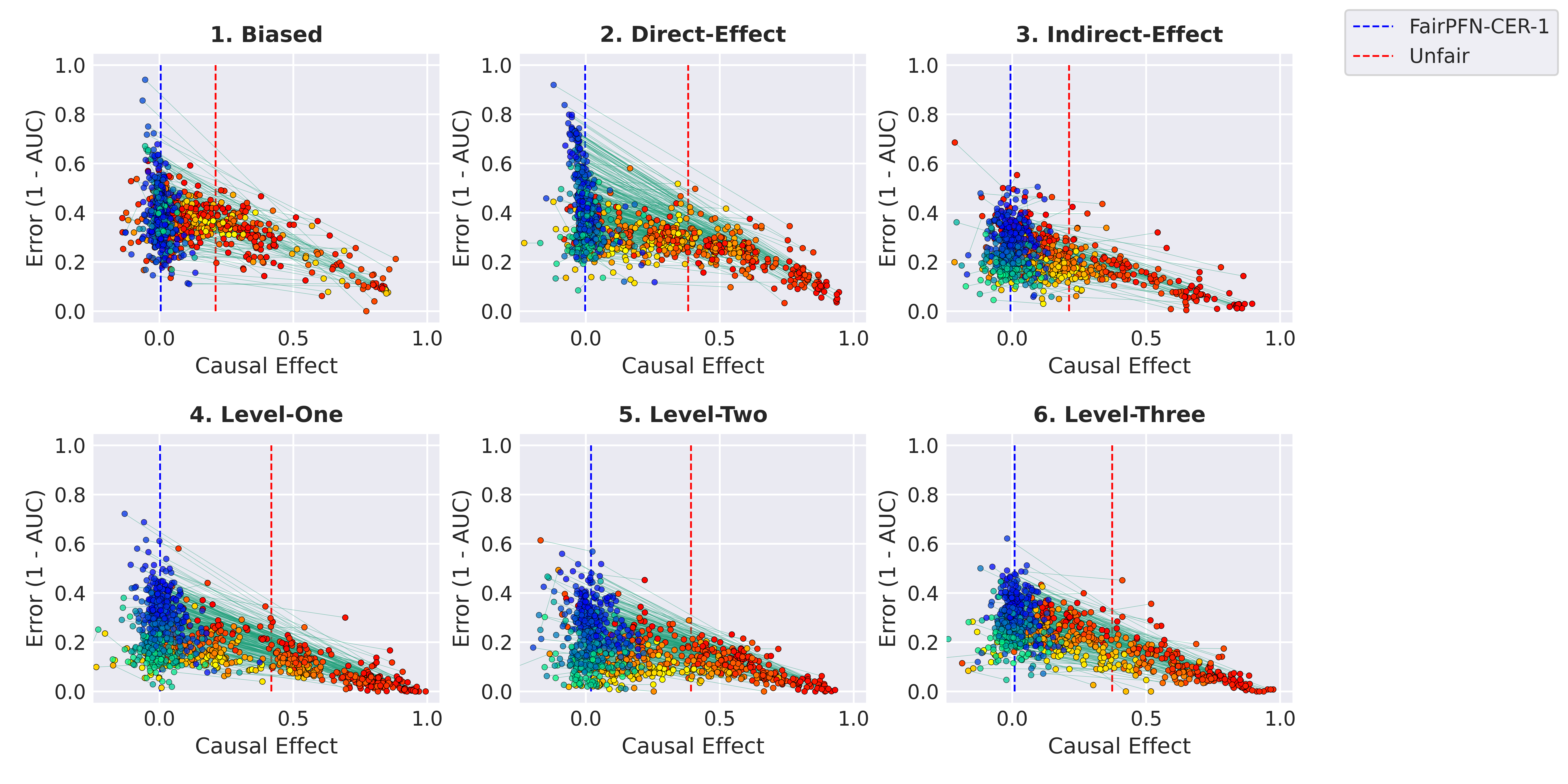}
    \caption{\textbf{Effect of Noise Terms (Synthetic)}: Causal Effect (TCE) and erorr (1-AUC) of FairPFN compared to the \texttt{Unfair} baseline on each individual dataset from our causal case studies. We provide a color gradient for both baselines (blue to green and red to yellow) to depict increasing amount of noise in the data. FairPFN consistently reduces the TCE on all benchmark groups, achieving lower error on datasets with larger amounts of noise.}
    \label{fig:compare}
\end{figure}

\paragraph{Synthetic Data} First, we evaluate FairPFN on our synthetic causal case studies, by visualizing the change in causal effect (DE, IE, or TE) before and after bias-mitigation with FairPFN (Figure \ref{fig:compare}), with a color gradient of blue to green to represent the increasing amount of noise in each dataset. We observe across all case studies that FairPFN learns to remove the causal effect of the protected attribute with a small variance and highlight two interesting effects. 

First, we observe on 5 out of 6 case studies that datasets with higher noise levels can generally be solved while maintaining a lower level of error. This could be due to 1) the lower \texttt{Unfair} TCE in these datasets or 2) the increased identifiability of SCMs with noise and non-linearity \cite{peters2014causal}. Additionally, we find that on the \texttt{Biased} case study, FairPFN often achieves an error (1-AUC) less than 0.5. This suggests that FairPFN does not revert to a random classifier when data is only causally influenced by protected attributes as there is still fair information (namely $\epsilon_{X}$ and $\epsilon_{y}$) in the data. Instead, FairPFN removes only the causal effect $w_{A}A^2$ in the corresponding structural equation, allowing the noise terms $\epsilon_{X}$ and $\epsilon_{y}$ to influence its predictions.

We also observe in Figure \ref{fig:effect_removal} that FairPFN dominates \texttt{EGR} in 5 out of 6 case studies, is on the Pareto Front in all 6, and always improves in terms of predictive performance compared to \texttt{CFP}. This is likely attributed to the effect observed in Figure \ref{fig:effect_removal} on the \texttt{Biased} case study, where FairPFN learns to remove only the causal effect of the protected attribute, still allowing all remaining information to influence its predictions.

\begin{figure}[h!]
    \centering
    \includegraphics[width=0.37\linewidth]{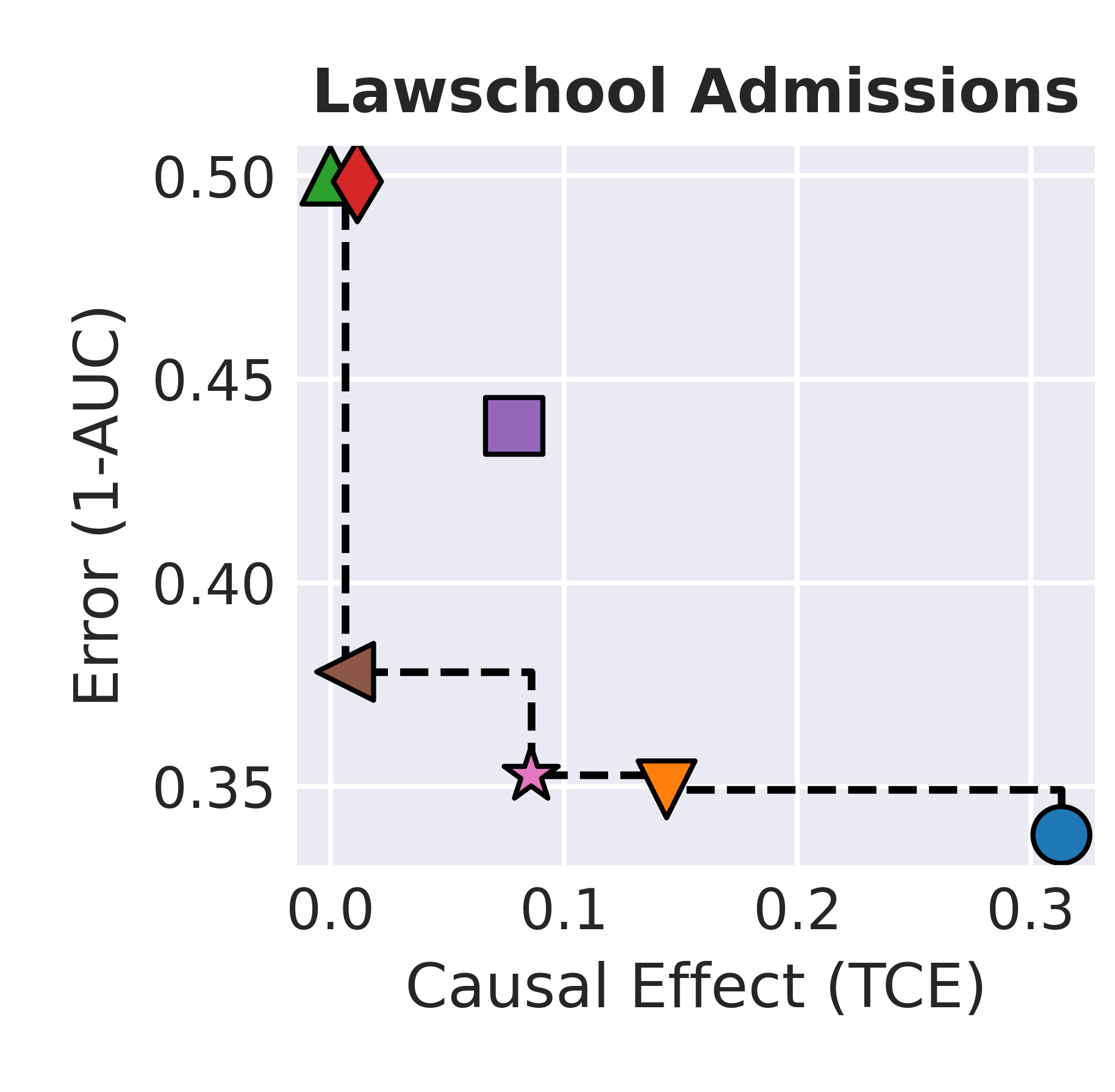}
    \includegraphics[width=0.61\linewidth]{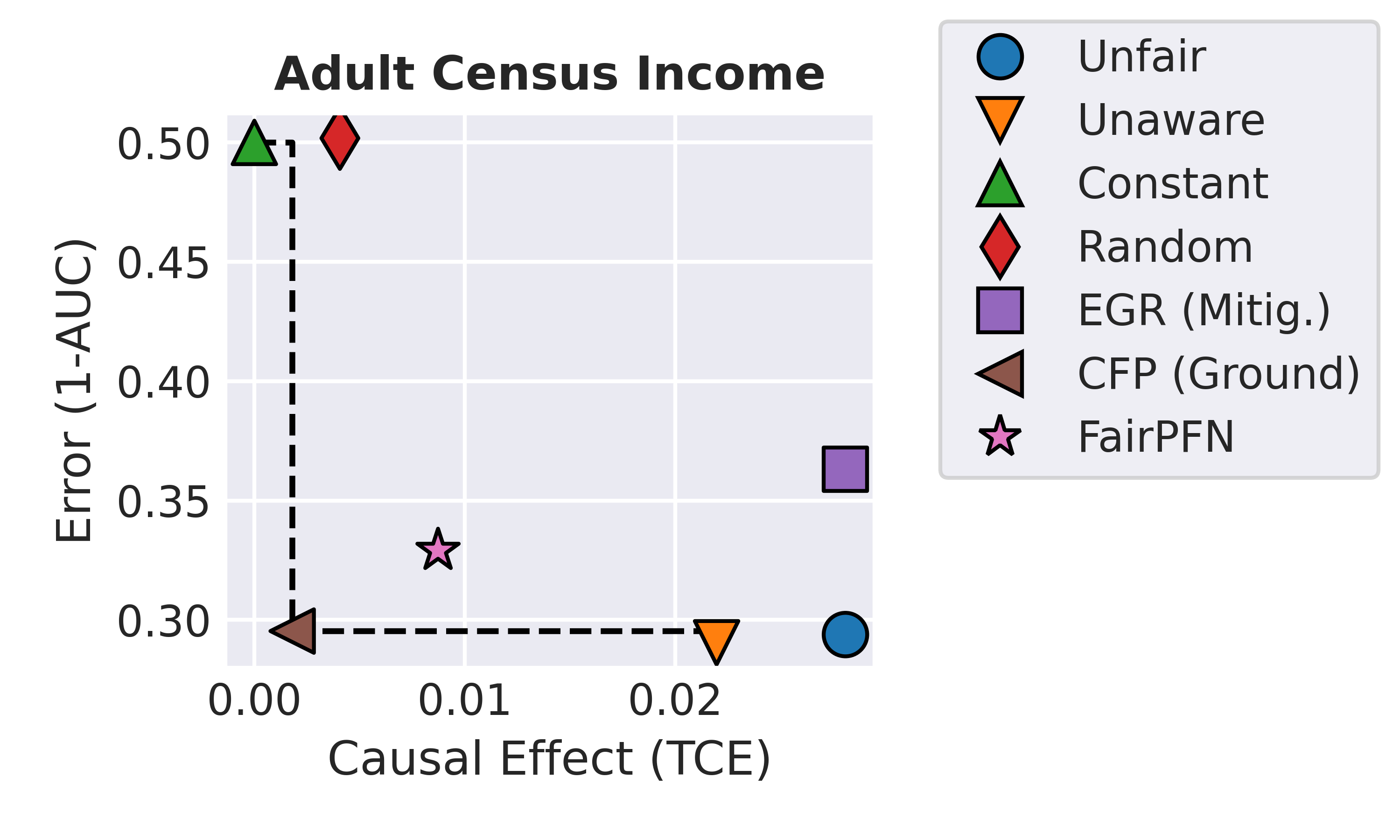}
    \caption{\textbf{Causal Effect Removal (Real-World):} Causal effect (TCE) and error (1-AUC) of FairPFN on our real-world datasets compared to other baselines. FairPFN is Pareto Optimal in both cases and provides a strong balance of causal fairness and accuracy.}
    \label{fig:mitig}
\end{figure}

\paragraph{Real-World Data}

We also evaluate FairPFN on the Law School Admissions and Adult Census Income datasets, using causal models fit to the structures posed in Figure to measure the TeE and MAE \ref{fig:real-world}. We note again that in evaluation FairPFN receives no information about the causal graphs or models.
In Figure \ref{fig:mitig}, we measure the causal effect across different baselines, observing that FairPFN shows significant improvement in terms of TCE compared to the \texttt{Unfair} and \texttt{Unaware} baselines. It also demonstrates competative TCE and improved error on the Law School dataset compared to the \texttt{CFP} baselines  On the Adult dataset, FairPFN is outperformed by the \texttt{CFP} baseline, which achieves dominating TCE and error. This outcome is likely explained by the fact that the \texttt{Unfair} TCE on the Adult dataset is already quite small (0.03), and thus the four fair noise terms in Figure \ref{fig:real-world} have a relatively higher representative capacity than in the Law School problem. However, FairPFN still reduces the TCE to less than 0.01, a very acceptable outcome in the broader scope of the problem.

\begin{figure}[h!]
    \centering
    \includegraphics[width=0.49\linewidth]{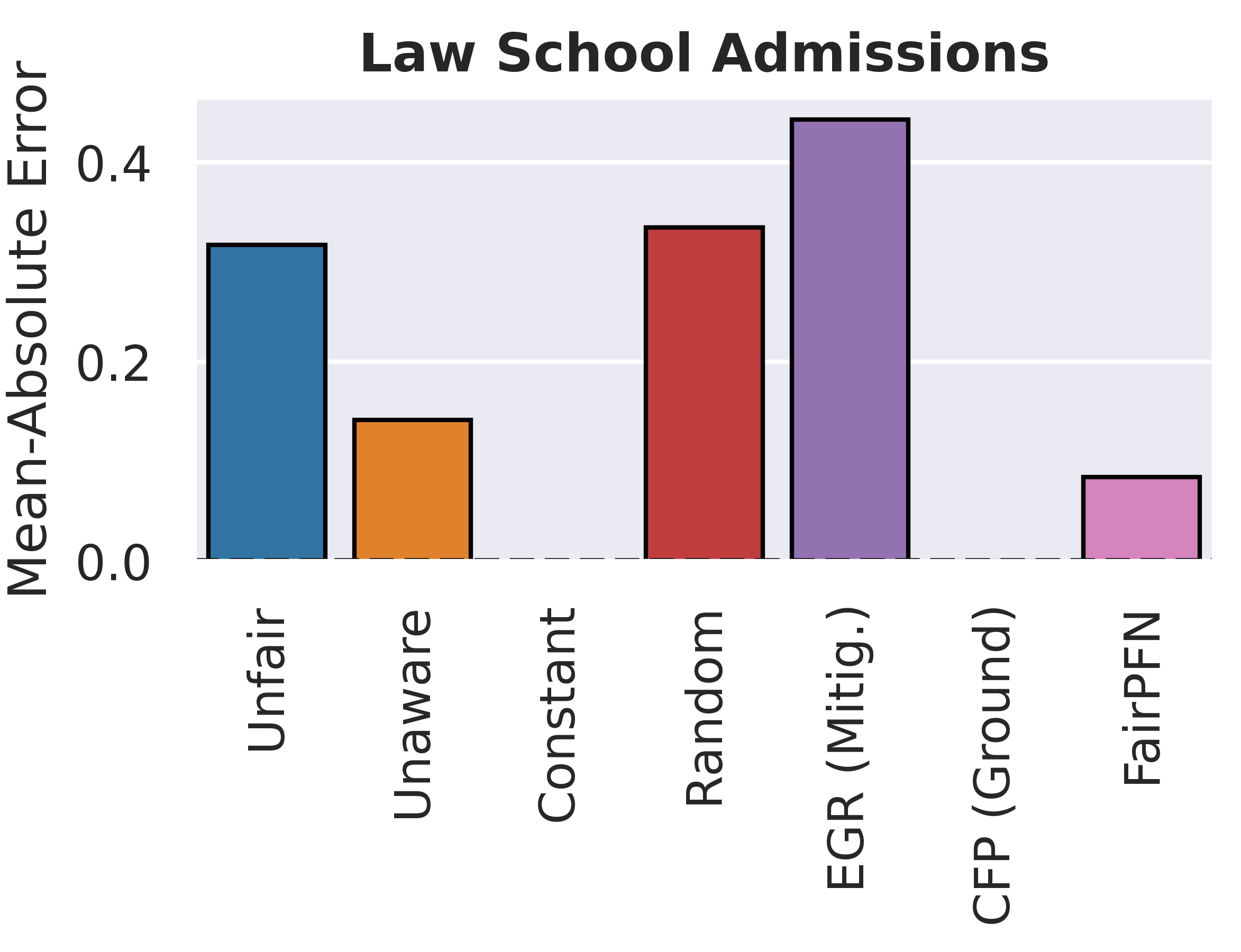}
    \includegraphics[width=0.485\linewidth]{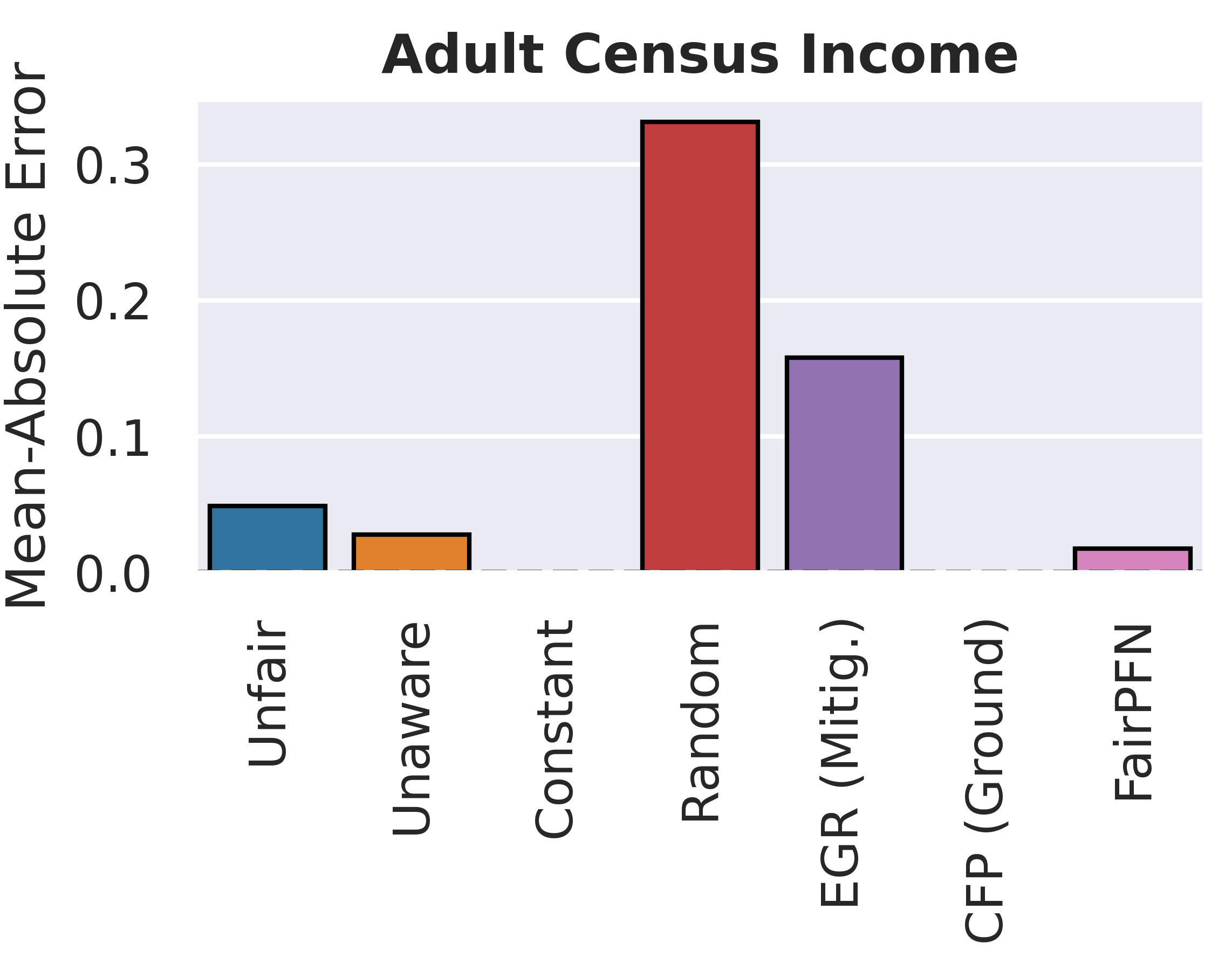}
    \caption{\textbf{Counterfactual Fairness (Real-World):} Mean Absolute Error (MAE) between predictive distributions on the original and counterfactual versions of our real-world datasets. FairPFN achieves competitive MAE with \texttt{CFP} and \texttt{Constant} baselines without having prior knowledge of the causal graph.}
    \label{fig:cntf}
\end{figure}

In Figure \ref{fig:cntf}, we also measure the MAE between the predictive distributions on the real and counterfactual datasets, $\hat{Y}_{real}$ and $\hat{Y}_{a \rightarrow a'}$. We observe that FairPFN achieves competitive MAE with \texttt{CFP} in both scenarios, learning to make counterfactually fair predictions without having access to the causal model or graph. We note that interestingly, \texttt{EGR} performs similarly poorly to \texttt{Random} in both scenarios, aligning with the intuition that randomization is not a counterfactually fair strategy as individuals do not receive consistent outcomes in either the real or counterfactual worlds. 

\section{Future Work \& Discussion}
\label{sec:conclusion}

In this study, we introduce FairPFN, a novel bias-mitigation technique that learns a pre-trained transformer to remove the causal effect of protected attributes in fairness-aware binary classification problems from observational data alone. FairPFN addresses a key limitation in the causal fairness literature by eliminating the need for prior knowledge of the true causal graph in fairness datasets, making it easier for practitioners to apply counterfactual fairness to complex problems where the underlying causal model is unknown. This expands the scope and applicability of causal fairness techniques, enabling their use in a broader range of scenarios. Looking ahead, we believe that FairPFN opens the door to several promising avenues of research.

\textbf{Real-World Evaluation} 
A crucial next step in FairPFN would be to train a module to predict the effect of interventions on the protected attribute, producing counterfactual datasets to evaluate on. Doing so with FairPFN could hold advantages in robustness as compared to using causal discovery techniques such as \cite{lorch2022amortizedinferencecausalstructure}, since our pre-trained transformer integrates over the possible causal explanations for the data.

\textbf{Transparency and Interpretability} 
In cases where a causal graph or a subset of causal relationships are known, incorporating this domain knowledge as additional input to the transformer could enhance both FairPFN's human-centricity and performance. Additionally, a future direction could involve predicting the causal graphs that explain the data, adding an extra layer of interpretability. 

\textbf{Fairness Preprocessing} 
By modifying FairPFN's output to predict not only fair outcomes but also fair versions of observational variables, we can improve interpretability and transparency while allowing practitioners to use their preferred ML model during deployment. FairPFN could also be repurposed as a generative model to create fair training data, increasing the performance of the selected model.

\textbf{Business Necessity} 
Incorporating these business-necessity from \cite{plecko2022causal} variables into our fairness prior could enable specifying variables through which to allow the causal effect of the protected attribute. This extension is similar to path-specific counterfactual \cite{peters2014causal}, which would also open up many more application areas, such as medical diagnosis, where the social effects of protected attributes like sex should be removed, yet their biological effects must be preserved to provide individualized treatment.


\newpage

\section*{Acknowledgements}

The authors of this work would like to thank the reviewers and the the organization of the ICML Next Generation AI Safety Workshop for the opportunity to share our work and recieve feedback. We would like to additionally thank the Zuse School ELIZA Master's Scholarship Program for their financial and professional support of our main author.

\newpage

\bibliography{references}
\bibliographystyle{icml2024}

\newpage
\appendix
\onecolumn

\section{Baseline Models}
\label{sec:baselines}

To compare FairPFN to a diverse set of traditional, causal-fairness, and fairness-aware ML algorithms, we also implement several baselines which we summarize below:

\begin{itemize}
    \item \texttt{Unfair}: A \texttt{TabPFNClassifier} is fit the entire dataset $(X, A, y)$
    \item \texttt{Unaware}: A \texttt{TabPFNClassifier} is fit to non-protected attributes $(X, y)$
    \item \texttt{Constant}:  A "classifier" that always predicts the majority class
    \item \texttt{Random}:  A "classifier" that randomly predicts the target
    \item \texttt{Level-One}:  A \texttt{TabPFNClassifier} is fit to non-descendant observables of the protected attribute $(X_{fair}, y)$ if any exist
    \item \texttt{Level-Two}: A \texttt{TabPFNClassifier} is fit to non-descendant unobservables of the protected attribute $(U_{fair}, y)$ if any exist
    \item \texttt{Level-Three}: A \texttt{TabPFNClassifier} is fit to noise terms of observables $(\epsilon, y)$ if any exist
    \item \texttt{EGR}: Exponentiated Gradient Reduction (EGR) for fairness metric DP as proposed by \cite{agarwal2018reductions}
\end{itemize}

We note that these baselines are specifically designed to provide ground truths of the best and worst that can be done in terms of fairness metrics and that certain baselines are only applicable to certain datasets. For example \texttt{Unfair}, \texttt{Unaware}, \texttt{Random}, \texttt{Constant}, and \texttt{EGR} are applicable on all synthetic and real-world datasets. \texttt{Level-One} is only applicable to  \texttt{Direct Effect}, \texttt{Indirect Effect} synthetic causal case studies. \texttt{Level-Two} is additionally applicable to the \texttt{Level-Two} synthetic case study, and \texttt{Level-Three} is additionally applicable to the \texttt{Level-Three} synthetic case study as well as the real-world datasets where the causal model is known and noise terms $\epsilon$ can be estimated.

\newpage

\begin{figure}[h!]
    \centering
    \includegraphics[width=\linewidth]{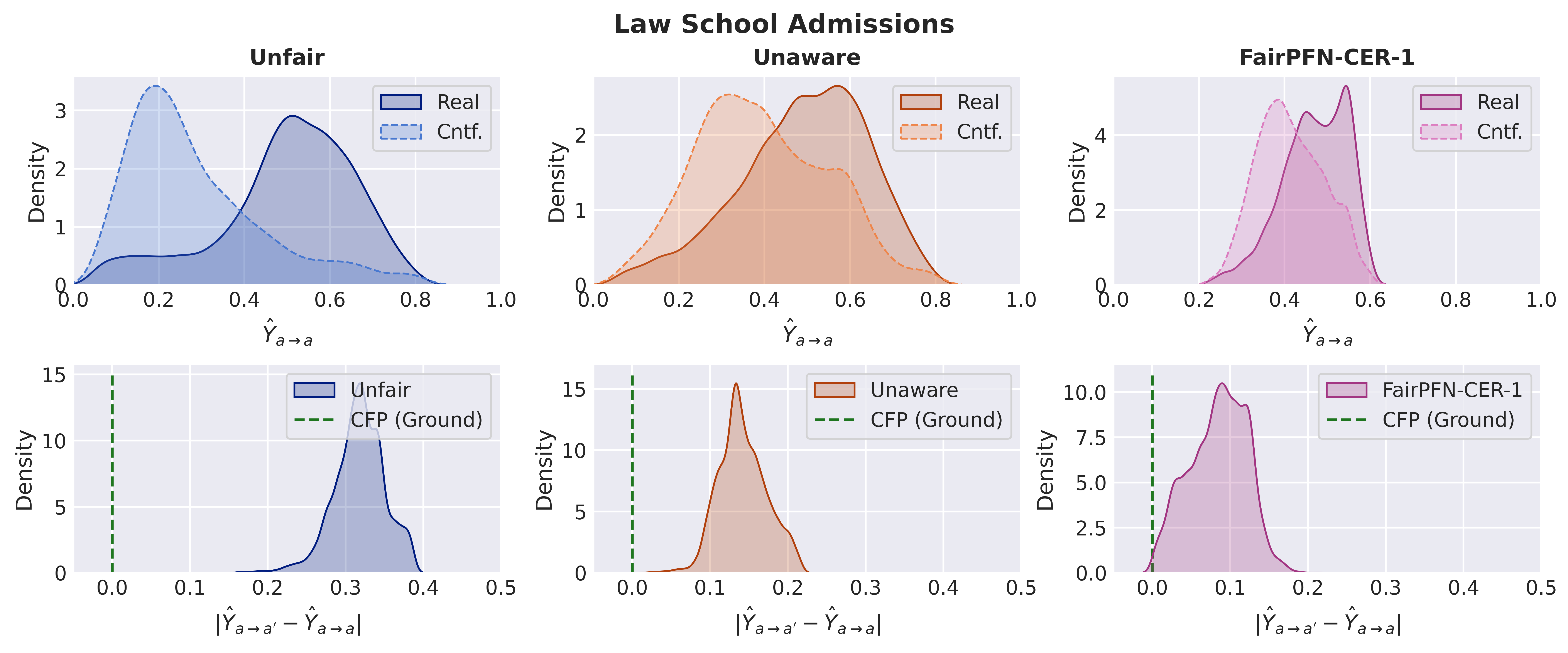}
    \caption{\textbf{Aligning Counterfactual Distributions (Law School):} Alignment of real and counterfactual predictive distributions $\hat{Y}$ and $\hat{Y}_{a \rightarrow a'}$ on the Law School Admissions problem. FairPFN best aligns the predictive distributions (top) and achieves the lowest mean (0.1) and maximum (0.2) absolute error.}
    \label{fig:enter-label}
\end{figure}

\begin{figure}[h!]
    \centering
    \includegraphics[width=\linewidth]{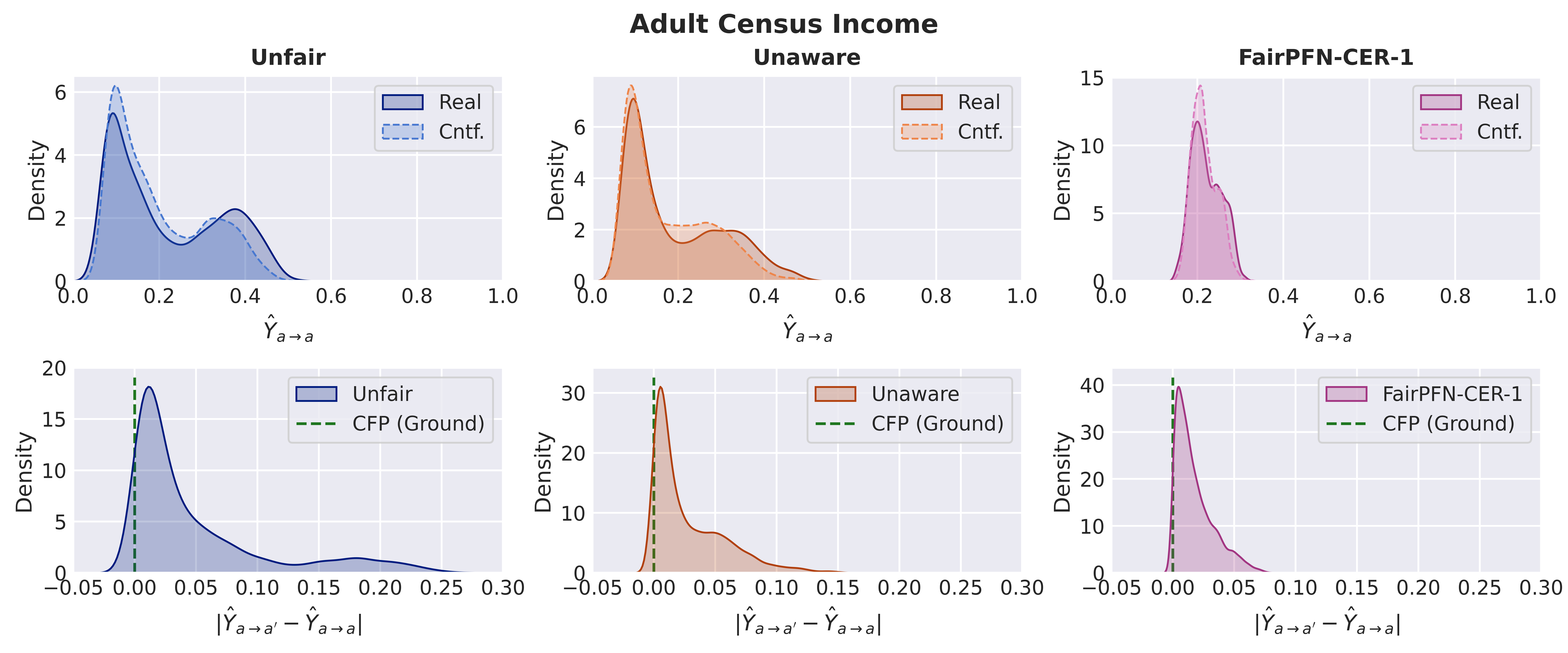}
    \caption{\textbf{Aligning Counterfactual Distributions (Adult):} Alignment of real and counterfactual predictive distributions $\hat{Y}$ and $\hat{Y}_{a \rightarrow a'}$ on the Adult Census Income problem. FairPFN best aligns the predictive distributions (top) and achieves the lowest mean (0.01) and maximum (0.75) absolute error.}
    \label{fig:enter-label}
\end{figure}

\begin{figure}[h!]
    \centering
    \includegraphics[width=\linewidth]{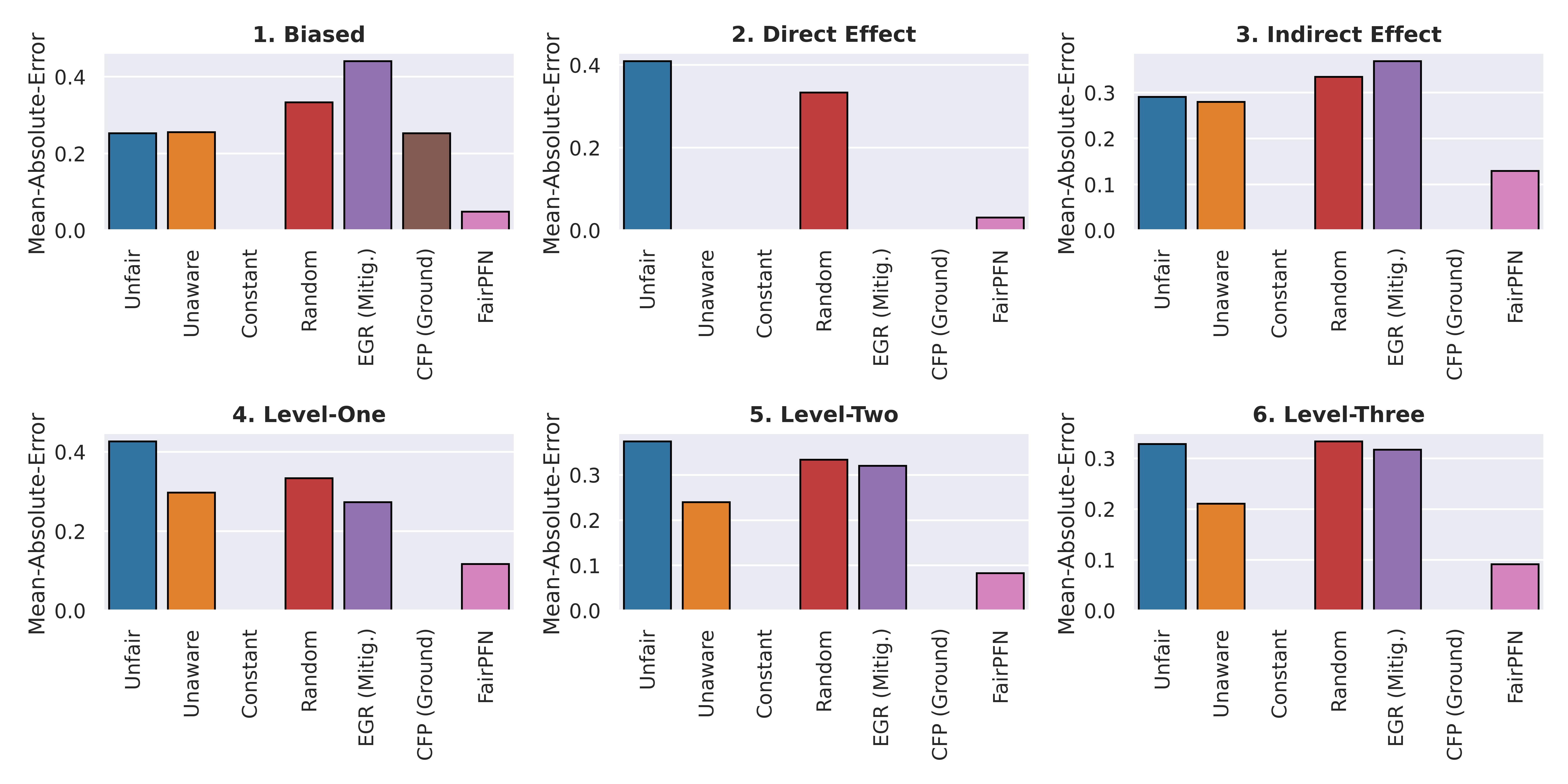}
    \caption{\textbf{Counterfactual Fairness (Synthetic):} Mean Absolute Error (MAE) between predictive distributions on the original and counterfactual versions of our causal case studies. FairPFN achieves competitive MAE with \texttt{CFP} and \texttt{Constant} baselines without having prior knowledge of the causal graph.}
    \label{fig:enter-label}
\end{figure}

\begin{figure}[h!]
    \centering
    \includegraphics[width=0.875\linewidth]{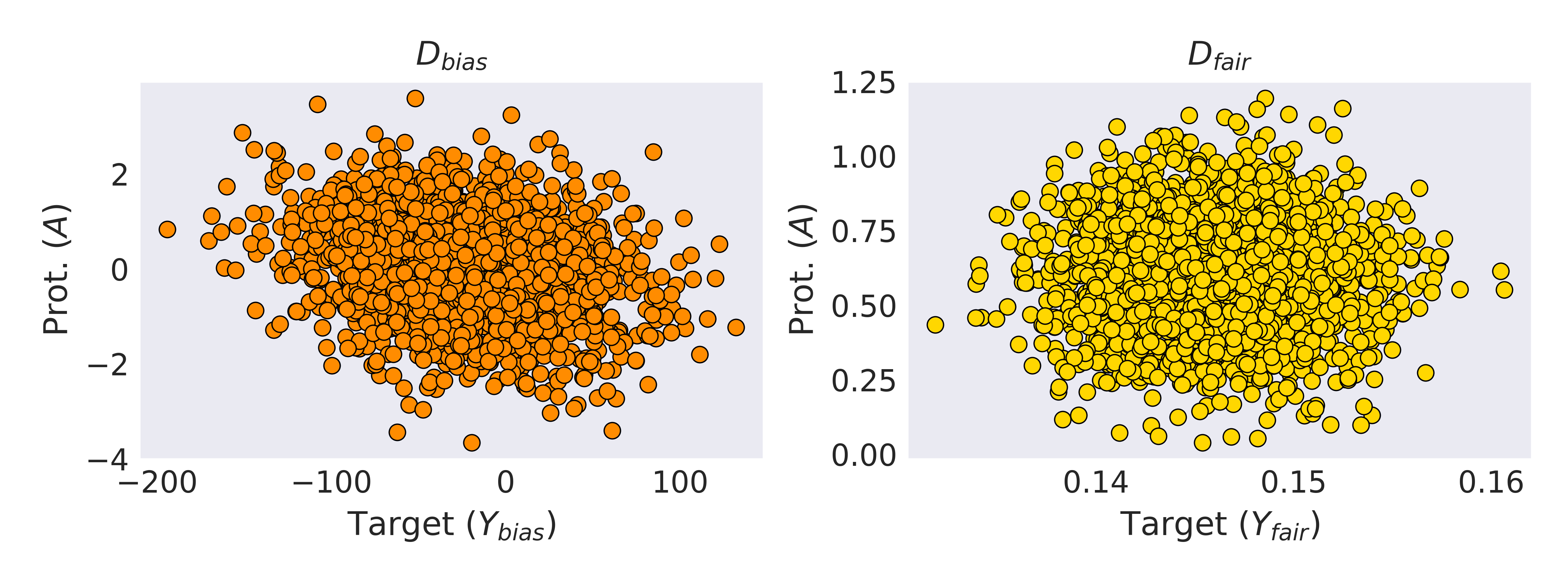}
    \caption{\textbf{Effect of Dropout:} Visualization of the effect of dropout on the outgoing edges of a protected attribute in a sampled MLP. In the biased dataset (left), the protected attribute has a slight negative correlation with the target, while in the fair dataset this effect is reduced to Gaussian Noise.}
    \label{fig:corr_removal}
\end{figure}

\begin{figure}[h!]
    \centering
    \begin{subfigure}
        \centering
        \includegraphics[width=0.37\linewidth]{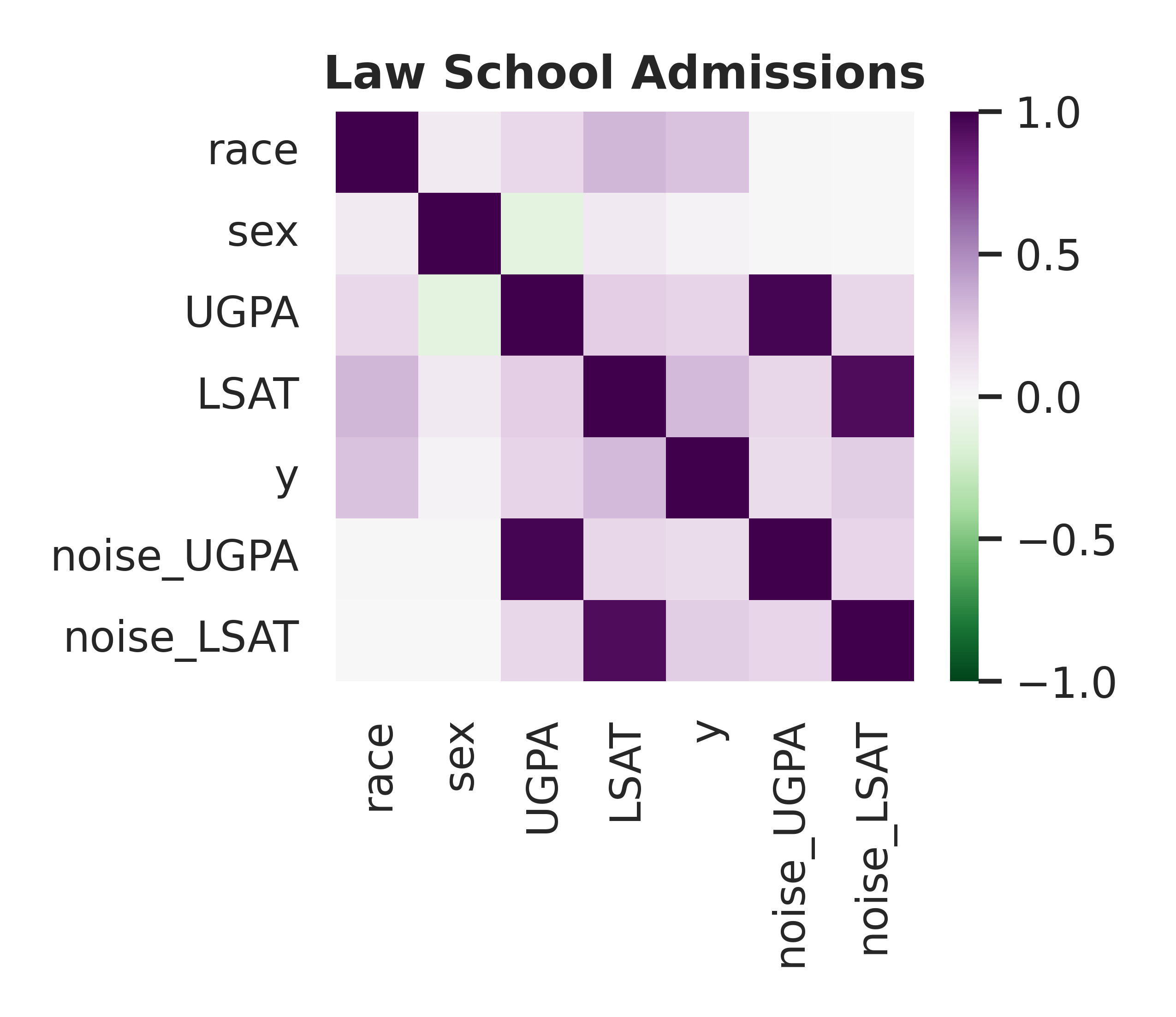}
        \label{fig:law_corr}
    \end{subfigure}
    \hfill 
    \begin{subfigure}
        \centering
        \includegraphics[width=0.59\linewidth]{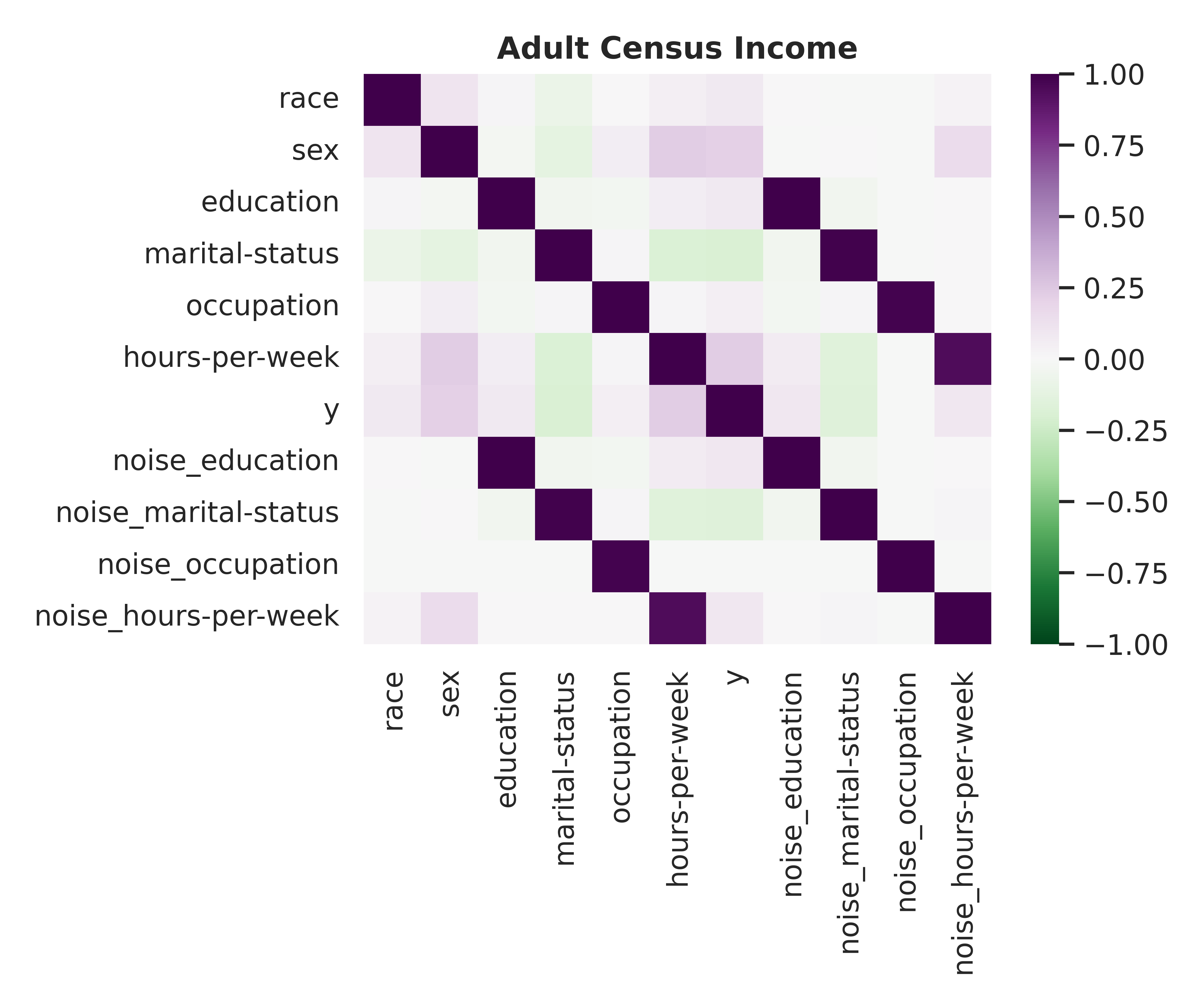}
        \label{fig:adult_corr}
    \end{subfigure}
    \caption{\textbf{Derivation of Noise Variables}: Pearson correlation of features including noise terms calculated using inverse probabilistic programming in \texttt{dowhy}'s \texttt{compute\_noise} functionality. Noise terms are uncorrelated with protected attributes and highly correlated with their corresponding observable.}
    \label{fig:noise}
\end{figure}

\end{document}